\pdfoutput=1
\documentclass{article}
\usepackage{arxiv}

\usepackage[bookmarks=false]{hyperref}      
\usepackage{url}            
\usepackage{booktabs}       
\usepackage{amsmath}
\usepackage{amsfonts}
\usepackage{bbold}
\usepackage{lipsum}
\usepackage{graphicx}
\usepackage{graphics}
\usepackage{pdfpages}
\usepackage{stmaryrd}
\usepackage[hypcap=false]{caption}

\title{Text-to-Image Generation with Attention Based Recurrent Neural Networks}

\author{
  \And \And \And
  Tehseen Zia\thanks{Corresponding author: \href{mailto:tehseen_zia@yahoo.com}{tehseen\_zia@yahoo.com}}\\
  \And 
  Shahan Arif\\
  \And
  Shakeeb Murtaza\\
  \And 
  Mirza Ahsan Ullah\\
  \And \And \And
}

\newcommand{\affiliation}{COMSATS University Islamabad, Islamabad Pakistan}

\begin{document}

\maketitle

\begin{abstract}
Conditional image modeling based on textual descriptions is a relatively new domain in unsupervised learning. Previous approaches use a latent variable model and generative adversarial networks. While the formers are approximated by using variational auto-encoders and rely on the intractable inference that can hamper their performance, the latter is unstable to train due to Nash equilibrium based objective function. We develop a tractable and stable caption-based image generation model. The model uses an attention-based encoder to learn word-to-pixel dependencies. A conditional autoregressive based decoder is used for learning pixel-to-pixel dependencies and generating images. Experimentations are performed on Microsoft COCO, and MNIST-with-captions datasets and performance is evaluated by using the Structural Similarity Index. Results show that the proposed model performs better than contemporary approaches and generate better quality images.  

\end{abstract}


\section{Introduction}
Generative modeling of natural images remains a fundamental image understanding problem. The capacity to handle the abundance of highly dimensional and highly structured imagery data allowed deep neural networks (DNNs) to achieve significant advances in generative image modeling \cite{1,2}. However, most of the contemporary image generation models are restricted to unconditional \cite{2,3,4,5,6} or class-based conditional image modeling \cite{7,8}. Nevertheless, since natural images are often accompanied with natural language descriptions (a.k.a. captions), incorporating this information could be valuable to improve generative capacity (i.e., complexity and expressivity) of models \cite{9,10,11}. Moreover, the leverage to generate images from captions may lead to better understanding of the generalized performance of the generative models as new scenes can be generated from novel text descriptions that are not shown during training \cite{10}. Furthermore, caption-based generative models could be used for building many practical applications, such as planning in virtual environments \cite{12} and neural artwork \cite{13}, etc.

Although numerous studies are recently conducted for learning generative models over both imagery and textual data, majority work is focused on learning to generate captions from images \cite{14,15,16}. On the contrary, generating images from captions is an opposite task of caption generation where the objective is to learn a model to predict images from captions. Caption-based image generation requires dealing with both language modeling and image generation which make it more challenging than caption generation.

One of the most critical objectives in the generative modeling is constructing models that have more capacity but are also tractable \cite{5,6}. One elegant approach to build such models is casting joint distribution modeling problem into a sequence of conditional distributions and learn models to predict the next pixel given all the previous pixels \cite{5,6,7}. Recurrent neural networks (RNN) are shown to be powerful models for this approach because of their capacity to model highly non-linear and long-range dependencies with shared parameterization over the sequence of conditional distributions \cite{5,6,17}.

The caption-based image generation models include stochastic latent variable based models \cite{10} and generative adversarial networks \cite{18}. While the formers are approximated by using variational auto-encoders and relied on the intractable inference that can hamper their performance \cite{19,20}, the latter are unstable to train due to their objective function that requires Nash equilibrium \cite{8,18}. The goal of this work is to illustrate how we can build a tractable and stable caption-based conditional image model over natural image space by using an autoregressive approach. In this regard, the contemporary approaches are either unconditional or class based conditional. In a recent study, an autoregressive based model is used for modeling images conditioning on text and spatial structure \cite{11}. This model relied on a third-party segmentation method and hence not an end-to-end learning method. The novelty of this work is to propose an attention-based framework to avoid explicit use of object/class detector by enabling the model to learn latent alignments between text description and pixels. This development enables the existing caption based autoregressive image models to surpass “objectness” and learn to generate images while attending to abstract concepts.

The work is primarily inspired by recent developments in neural machine translation and image caption generation with an attention-based mechanism. The main idea is to use the encoder-decoder framework where an attention-based encoder is used to learn word-to-pixel dependencies and a conditional autoregressive based decoder for learning pixel-to-pixel dependencies within images. The key contribution of this work is to introduce an attention-based autoregressive image model for learning to generate images from text descriptions. Also, we empirically validate the efficacy of the proposed model by achieving superior performance on MS-COCO and MINIST datasets. The remaining paper is organized as related work is discussed in Section 2, the background is described in Section 3, the proposed model is presented in Section 4, the experimental setup is given in Section 5, results are presented in Section 6, and finally, the paper is concluded in Section 7. 

\section{Related work}
Unlike the discriminative models where significant developments have been achieved with deep neural networks in various tasks including image recognition \cite{21}, speech transcription \cite{22} and machine translation \cite{23}, generative models have not yet gained the same level of appreciations. The earliest work has mostly concentrated on devising Deep Belief networks \cite{24} and advancing Boltzmann Machines \cite{25}. Though the models are exquisite in their own merits, the higher computational cost makes them infeasible to scale on larger datasets. To cope with this issue, Variational Auto-encoder (VAE) (i.e., a neural network with stochastic latent variable) is introduced in \cite{19,20}. The model is an encoder-decoder framework; the encoder is employed to approximate a posterior distribution and decoder is involved in generating the data stochastically from the latent variable. An extension of VAE is introduced in \cite{4} and referred to as Deep Recurrent Attention Writer (DRAW). The additional feature of the DRAW was the integration of a novel attention mechanism into the VAE model. The unconditional DRAW model was further extended in \cite{10} with the ability to model conditional distribution. The resultant alignDRAW model enabled the generation of natural images from natural language descriptions. An essential characteristic of the model is to interface image generative modeling with language modeling which is achieved by conditioning previously unconditional latent variable. Because the VAE models are approximated intractably by using latent variables, a group of researchers is focused on developing tractable generative models. 

An elegant approach for tractably modeling the joint distribution is casting it as a sequence of conditional distributions and hence turning it into sequence modeling problem \cite{6}. This approach is used in autoregressive models such as NADE \cite{26} and fully-visible neural network architectures \cite{27}. An extension of the method for modeling two-dimensional grayscale images and textures are presented in \cite{6} and referred to as recurrent image density estimator (RIDE). Two-dimensional long short-term memory (LSTM) \cite{28} (a variant of RNNs for capturing long-term dependencies) are used as an auto-regressive model to capture pixel dependencies from left-to-right and top-to-bottom. While RIDE estimates continuous distribution over pixel values, it is shown that discrete modeling distribution over pixel value leads to achieve better results \cite{5}. The resultant model architecture, well-known as the pixelRNN model, is a deep two-dimensional LSTMs architecture (i.e., with stacked LSTM layers) that includes residual connections between layers. To faster the training process, a variant of pixelRNN, known as pixelCNN is also devised \cite{5}. The model allows parallel computations with bound dependency range in contrast to unbound dependency range in pixelRNN. The convolution masks are used to ensure the autoregressive density model remains causal; a pixel depends only on above and left pixels. The masked convolution is further applied to successively generate three color channels by conditioning red pixels on previous pixels (i.e., above and left), green on red and blue on red and green. Though pixelCNN has speed-up training of autoregressive models, they are costly at inference time. To deal with the issue, a parallelized pixelCNN is proposed in \cite{29} where the authors have made inference procedure faster by modeling certain pixel groups as conditionally independent. While the above autoregressive models deal with unconditional density estimation of nature images, this work has focused on conditional density estimation. To enable the pixelCNN model for learning conditional distribution, conditional pixelCNN is proposed in \cite{7}. The model is composed of several convolutional layers with shortcut connections to pass the output of each layer into the penultimate layer before the pixel logits. 

In \cite{11}, the pixelCNN is further enabled to generate images conditioning on text and spatial structure. The conditioning information is arranged according to their relevant spatial location in a feature map to preserve the location structure. The available segmentation of MS-COCO dataset for 80 classes is used for localizing the information. The textual information is encoded by using character-CNN-GRU and tiled spatially. The textual and location feature maps are concatenated and convolved by using dilated convolution before applying PixelCNN to generate the image. A key issue with the model is its reliance over a third-party segmentation method and hence not an end-to-end learning method. Secondly, the performance of the generation method depends on the performance of the segmentation method. In contrast to this model, we are proposing an attention-based framework to avoid explicit use of object/class detector by enabling the model for learning latent alignments between text description and pixels. This development enables the model to surpass “objectness” and learn to generate images while attending to abstract concepts. A similar approach is adopted in \cite{10} but with intractable DRAW decoder. 

Despite autoregressive models, Generative Adversarial Networks (GANs) are another kind of models that avoid intractable distribution estimating by using noise-contrastive estimation \cite{3}. The key idea is to use two networks; one for generating samples from a uniform distribution and second for discriminating between real and generated samples. In \cite{8,30}, GANs are enabled to model conditional distribution by conditioning both generator and discriminator on a given class label. The GANs are further extended in \cite{18} to generate images from captions and image spatial constraints (such as the location of human joint or bird fragments). Also, they have demonstrated that translation, stretching and shrinking of objects can be controlled through controlling key points of an image. While GANs have their own merits such as fast sample generation as it does not require generating images sequentially pixel-by-pixel, training GANs is unstable since it requires achieving Nash equilibrium of a game. Further, GANs do not provide the likelihood of the learned models which make it challenging to evaluate the efficacy of the generative models in an established and principled way. In comparison, the autoregressive models are much more straightforward and stable to train as well as provide the likelihood of the models. Therefore, empowering autoregressive models with controlled image generation is an important frontier. 

\section{Background}
In this section autoregressive image modeling is briefly described to better explain the proposed model structure latterly.
\subsection{Autoregressive Image Modeling with PixelRNN}
\begin{figure}[!h]
	\centering
	\includegraphics[clip,width=5.5cm]{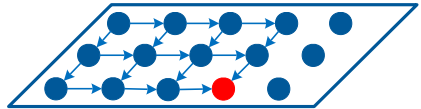}
	\captionsetup{justification=centering}
	\caption{Autoregressive Image Modeling}
	\label{fig:1}
\end{figure}

Consider an image $\mathbb{x}$ consisting of $n\times n$ pixels as a one-dimensional sequence of pixels $\mathbb{x}=(x_1,\ x_2,\ \ldots,\ x_{n^2})$ (e.g., formed by concatenating rows horizontally). This supposition allows fragmentation of joint distribution $p(\mathbb{x})$ into the product of the conditional distribution as:
\begin{equation}
    p\left(\mathbb{x}\right)=\ \prod_{i=1}^{n^2}{p(x_i|x_{1:i-1})}
\end{equation}
The distributions over pixel values can be modeled by using continuous distribution \cite{6} or discrete distribution \cite{5}. However, discrete distribution-based approach has recently gained in popularity because of its simple representation, ease of training, and better performance. In this approach, each conditional distribution is regarded as a multinomial and parameterized with a softmax layer. To model the product of conditional distributions, long short-term memory networks (LSTMs) \cite{28} or gated recurrent neural networks (GRU) \cite{31} based RNNs are commonly employed due to their ability to capture long-term dependencies. The contextual dependencies between pixels are captured by processing the images sequentially row-by-row from top-to-bottom \cite{5,6} as shown in Figure \ref{fig:1}.

\begin{figure}[!h]
	\centering
	\includegraphics[clip,width=5.5cm]{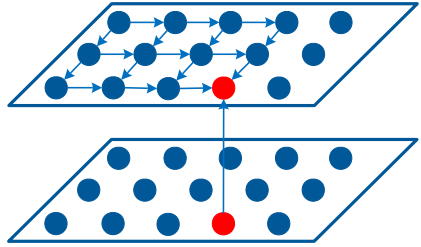}
	\captionsetup{justification=centering}
	\caption{Autoregressive Image Modeling using Recurrent Neural \\ Networks: input-to-state and state-to-state transactions.}
	\label{fig:2}
\end{figure}

The input pixels are processed with a one-dimensional convolution layer in order to obtain translation invariant features. In each step of LSTM, input-to-state and state-to-state transformations are computed and used to determine the values of four gates (output gate $o_i$, forgot gate $f_i$, input gate $i_i$ and content gate $g_i$) cell state $c_i$ and hidden state $h_i$ of LSTM cell as:
\begin{equation}
    \left[o_i,\ f_{i,\ }i_{i,\ }g_i\right]=\ \sigma(W^{ss}{\otimes}h_{i-1}+W^{is}{\otimes x}_i)
\end{equation}
\begin{equation}
    c_i=f_i\odot c_{i-1} + i_i \odot g_i
\end{equation}
\begin{equation}
    h_i=o_i\odot\tanh(c_i)
\end{equation}
Here $h_{i-1}$ and $c_{i-1}$ denotes previous hidden and cell states respectively each with a size of\ $h\times n\times1$. The symbols $\otimes$ and $\odot$ symbolizes the convolutional and element-wise multiplication operators. The weight matrices $W^{ss}$ and $W^{is}$ represents kernel weights for state-to-state and input-to-state transformations, respectively. The input-to-state and state-to-state computations in each step of autoregressive image modeling are illustrated in Figure \ref{fig:2} (a step corresponds to a pixel location). The lower layer symbolizes an image and upper layer represents an RNN layer (i.e., LSTM or GRU). The RNN performs computation at each pixel location based on the pixel input corresponding to the location and feedback from previous pixels (i.e., left and up) in the form of hidden states of previous RNN layers.

\section{Proposed Approach}
In this section, we present our proposed approach for text-to-image generation. As an image is regarded as a sequence of pixels in auto-regressive image modeling, text-to-image generation can be viewed as a sequence-to-sequence translation problem. An encoder-decoder framework (a.k.a. sequence-to-sequence model) has been a widely recognized approach for dealing with such translation problems in recent times. The encoder is employed to map input text description into annotation vectors as described in Section. The decoder is used to generate an output sequence from annotation vectors as explained in Section. A vital feature of the framework is to model the dependency between annotation vectors and pixels by using a soft attention-based mechanism. The model is shown to be very successful in neural machine translation \cite{23} and image captioning tasks \cite{14}.

\subsection{Encoder: Language Modeling with Bidirectional RNNs}
The proposed model takes a text description $y$ as input and generates an image $\mathbb{x}$ as an output. The language model process text description in terms of a sequence of words $y\ =\ (y_1,\ y_2,\ldots,\ y_N)$ encoded in standard 1-of-K words form, where $K$ and $N$ are respectively vocabulary size and sequence length of $y$. Bidirectional recurrent neural networks (BRNNs) are employed for representing each word $y_i$ into an m-dimensional feature vector ${\ h}_i^{lang};\ i=1,\ 2,\ \ldots,\ N$. In particular, two long short memory (LSTMs) networks are used to process the sequence of words in forward and backward directions to respectively compute a series of forward hidden states $[{\overrightarrow{h}_1^{lang}},\ {\overrightarrow{h}_2^{lang}},\ \ldots,\ {\overrightarrow{h}_N^{lang}}]$ and a series of backward hidden states $[\overleftarrow{h}_1^{lang},\overleftarrow{h}_2^{lang},\ldots,\overleftarrow{h}_N^{lang}]$. Finally, both the series are concatenated as $h_i^{lang}=[{\overrightarrow{h}_i^{lang}},{\overleftarrow{h}_i^{lang}}],1\leq i\geq N$ in order to attain a single series of hidden states $[h_1^{lang},\ h_2^{lang},\ \ldots,\ h_N^{lang}]$. 

\subsection{Decoder: Image Generation with Conditional Pixel-RNNs}
\begin{figure}[!h]
	\centering
	\includegraphics[clip,width=10cm]{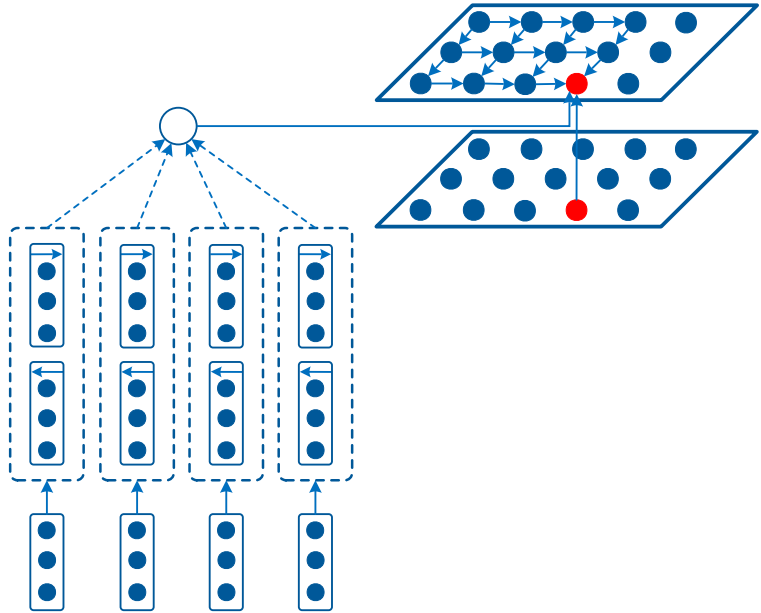}
	\captionsetup{justification=centering}
	\caption{Attention-based Autoregressive model for text-to-image generation.}
	\label{fig:3}
\end{figure}

To generate an image\footnote{To make notations simple, we assumed that the images $\mathbb{x}\in\mathbb{R}^{h\times w}$ have $h\times w$ size and consists of a single-color channel.}  $\mathbb{x}$ given the text description $y$, the Pixel-RNN network is extended to contain a text representation $h^{lang}$ at each step while generating a new pixel, as shown in Figure \ref{fig:3}. That is, unlike the original Pixel-RNN where the model is only conditioned on previous pixels, we additionally conditioned Pixel-RNN on a hidden state of a language model for modeling $p(\mathbb{x}|h^{lang})$: 
\begin{equation}
    p\left(\mathbb{x}\middle| h^{lang}\right)=\ \prod_{j=1}^{L}{p(x_j|x_1,\ \ldots,\ x_{j-1},\ h^{lang})}=gen\ (x_{j-1},\ h_j^{pix},\ c_j^{lang})
\end{equation}
Here gen represents conditional Pixel-RNN and $h_j^{pix}$ is hidden state of the conditional Pixel-RNN at time $j$, computed as: 
\begin{equation}
    h_j^{pix}=f(h_{j-1}^{pix},\ x_{j-1},\ c_j^{lang})
\end{equation}
It should be noticed that in this conditional Pixel-RNN the probability of a generating a pixel is conditioned on previous pixels and a distinct language context vector $c_j^{lang}$ for each pixel. The language context vector $c_j^{lang}$ is computed based on a sequence of hidden states $[h_1^{lang},\ h_2^{lang},\ \ldots,\ h_N^{lang}]$ to which an encoder maps the textual description, as defined in \cite{23}:
\begin{equation}
    c_j^{lang}=\sum_{i=1}^{N}\alpha_{ji}h_i^{lang}
\end{equation}
Each hidden state $h_i^{lang}$ contains information about the textual description with a key attention on the segments contiguous to the $i^{th}$ word of the description. The weight $\alpha_{ji}$ of every annotation vector $h_i^{lang}$ is obtained as: 
\begin{equation}
    \alpha_{ji}=\ \frac{\exp(e_{ji})}{\sum_{k=1}^{N}{\exp(e_{jk})}}
\end{equation}
where
\begin{equation}
    e_{ji}=a(h_{j-1}^{pix},\ h_i^{lang})
\end{equation}
The alignment model $a$ notches the matching of $i^{th}$ annotation $h_i^{lang}$ and $j^{th}$ pixel and implemented with feed-forward neural networks with $h_{j-1}^{pix}$ and $h_i^{lang}$ inputs and trained with the gradients of Pixel-RNN cost function. An interpretation of the context vector $c_j$ is that it is a computation of expected annotation with the expectation over all possible alignments. Hence, $\alpha_{ji}$ can be regarded as the probability that $j^{th}$ pixel is generated as a result of $i^{th}$ words of the description. In other words, Pixel-RNN decoder chooses segments of the description to pay attention to while generating each pixel in the image.

The proposed approach can be conceptualized as a reverse methodology of semantic segmentation. While a semantic segmentation maps subset of pixels to a class label (word); the proposed approach does the reverse by mapping the word(s) to subset of pixels. The proposed approach can also be considered as the reverse of image caption with attention mechanism \cite{14} because in this approach the model attends some pixels and generate a phrase, our model reversely attends a phrase of caption and generate a pixel.

\section{Experimental Setup}
\subsection{Datasets}
\begin{figure}[!h]
	\centering
	\includegraphics[clip,width=10cm]{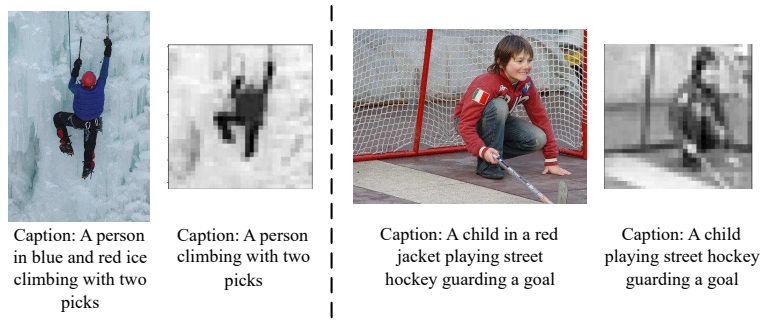}
	\captionsetup{justification=centering}
	\caption{Two examples from COCO dataset. Each example contains an original image with caption and pre-processing image with caption used in training.}
	\label{fig:4}
\end{figure}

In order to perform experiments with our proposed model, we employ Microsoft COCO \cite{32} and MNIST datasets. The COCO is a big dataset of natural images comprising of 82,783 images with no less than five captions per image. The dataset poses generative modeling a challenging task due to the massive diversity of objects, styles, and backgrounds among images. We considered the first five captions for each image in case the captions are more than five, in order to be consistent with the previous works. Also, for the sake of uniformity with other small image datasets, the sizes of the images are rescaled to 32-by-32. Further, the images are converted to gray scale for faster training of the proposed network. This preprocessing is performed by using the OpenCV library.  The color information is also excluded from the captions accordingly by creating a dictionary of colors and then captions are processed to discard the color information. This pre-processing is carried out using nltk libraries. Examples of the preprocessed dataset are given in Figure \ref{fig:4}.

The experiments on MNIST dataset are performed by using MNIST-with-captions dataset as used in \cite{10}. The images in the dataset\footnote{MNIST-with-caption dataset is available at \url{https://github.com/mansimov/text2image}. The dataset is created by using same code presented on this link.} are synthetically created by arranging either one or two digits horizontally or vertically in a non-overlapping manner over a 60-by-60 canvas of blank image. The captions of these images are also synthetically generated as “The digit 3 is at the bottom of the digit 0”. Examples of the dataset are given below in Figure \ref{fig:5}.

\begin{figure}[!h]
	\centering
	\includegraphics[clip,width=7cm]{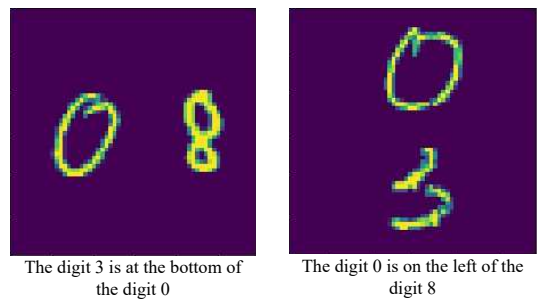}
	\captionsetup{justification=centering}
	\caption{Two examples from MNIST-with-captions dataset}
	\label{fig:5}
\end{figure}

\section{Training and evaluation}
The proposed alignPixelRNN model is trained with weighted cross-entropy cost function of a sequence of logits and sequence of the raw pixel value. To overcome the gradient exploding/vanishing problem, the clipping method is used where the gradient is clipped at 1.0. The parameter updating is performed by using RMSProp algorithm as recommended in \cite{5} with a default learning rate (i.e., 0.001). A small batch size of 16 is used as it seems to regularize the models \cite{10} and batches are generated by using a random sampling method. The model is implemented in Tensorflow toolkit \cite{33} by using a sequence-to-sequence wrapper with a caption as encoder input and image as decoder output. The sequence length of encoder’s input is kept as a variable due to varying lengths of captions. The sequence length of the decoder’s output is 1024 (32 x 32). An embedding layer with 512 neurons is used as the first layer of the encoder for word-to-vector representations, and each LSTM layer consists of 512 neurons. As described above, word-to-pixel dependencies are modeled by using an attention-based mechanism. In this regards, we employed Luong attention based mechanism as used in the baseline AlignDraw model.  We noticed that beam search based conditional image generator is much faster and equivalent in performance with greedy search-based decoders and used it for further evaluations. Finally, the model is trained on GPUs.

\section{Results}
\begin{figure}[!h]
	\centering
	\includegraphics[clip,width=9cm]{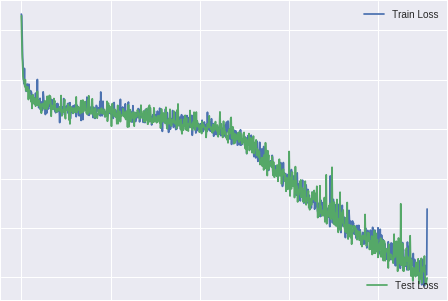}
	\captionsetup{justification=centering}
	\caption{Train test loss curves}
	\label{fig:6}
\end{figure}

The qualitative performance evaluation of the proposed model in comparison with previous models is shown in Table 1. We closely followed the experimentation and evaluation procedure of \cite{10} and ranked the test images conditioned on the captions on the basis of the variational lower bound of the log-probabilities. The recall measure is used to evaluate the quality of the generative model. Next, we gauged the performance of the model by used the Structural Similarity Index measure (SSIM) which computes the similarity between two images \cite{34}. The SSIM employs contrast masking, luminance, and inter-dependencies of nearer pixels for calculating the similarity between images. The computation is performed on two image patches (a,b) of common size $(N\times N)$ as:
\begin{equation}
    SSIM\left(a,b\right)=\frac{(2\mu_a\mu_b+c_1)(2\sigma_{ab}+c_2)}{(\mu_a^2+\mu_b^2+c_1)(\sigma_a^2+\sigma_b^2+c_2)}
\end{equation}
Where $\mu_a$ and $\mu_b$ are averages of patches a and b respectively, $\sigma_a^2$ and $\sigma_b^2$ are variances of patches a and b respectively, $\sigma_{ab}$ is covariance of  $a$ and $b, c_1$ and $c_2$ are variables for stabilizing the weak denominator. We generated fifty images for each test caption in the dataset for calculating the SSI on test dataset by following the work of \cite{10}. The models that we compared with consists of variational models including Fully-Connected (fully-conn) Variational Autoencoder (VAE) and Convolutional Deconvolutional (Conv-Deconv) VAE, as well as DRAW models comprising of SkipthoughtDRAW \cite{35}, noalignDRAW and alignDRAW \cite{10}. It can be seen from Table \ref{tab:1} that the autoregressive models outperformed variational and DRAW models in both recall and SSI metrics. We believe this is due to the reason that the proposed model does not overfit and better scale to the training data. This can be observed from the train-test loss function as shown in Figure \ref{fig:6}.

\begin{figure}[!t]
	\centering
	\includegraphics[clip,width=15cm]{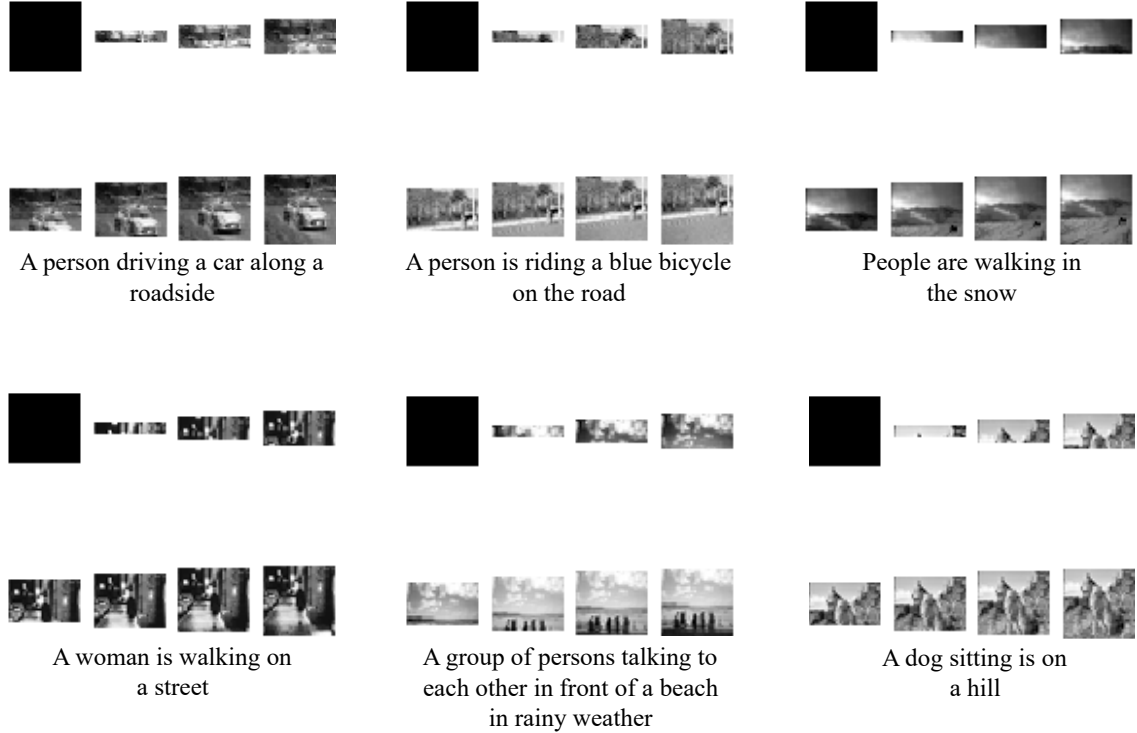}
	\captionsetup{justification=centering}
	\caption{Some fine generated examples of the alignPixelRNN model.}
	\label{fig:7}
\end{figure}

Some samples images generated by the model are shown in Figure \ref{fig:7}. The images from left-to-right show the output of the model after a quarter of iterations, half of the iterations and final iteration. It can be seen that the generated images are small and grayscale which is a reflection of constraints on the model size due to computational complexity and GPU memory requirements. However, image enhancement and colorization have been remarkably developed in recent times with readily available models that can be used to enhance and colorize the generated images of our model. Unfortunately, this may not be a flawless solution because of the fact that the entire model was not learned in an end-to-end manner. Thus, future work would be required to find a model structure that can avoid the need for separate post-processing step and uplift the computational and memory constraints. Nevertheless, it can be seen that the proposed model improves the visual quality and understandability of the generated images. A key differentiation between the generated images of baseline alignDRAW and proposed alignPixelRNN is that while alignDRAW generated an image at once and refined it iteratively, alignPixelRNN generate image iteratively.  Also, the generated images of alignPixelRNN are seemed to be sharper than the images of alignDRAW.

\begin{table}[!h]
\centering
\caption{Results of proposed (i.e., PixelRNN) in comparison with related models on MS COCO dataset. R@K is recall@K (higher is better). SSI is the Structural Similarity Index (higher is better). Earlier results are taken from \cite{10}.}
\label{tab:1}
\begin{tabular}{lccccc}
\hline
\multicolumn{1}{c}{\textbf{Model}} & \textbf{R@1} & \textbf{R@5} & \textbf{R@10} & \textbf{R@50} & \textbf{Image Similarity (SSI)} \\ \hline
Fully-Conn VAE & 1 & 6.6 & 12 & 53.4 & 0.156 $\pm$ 0.10 \\
Conv-Deconv VAE & 1 & 6.5 & 12 & 52.9 & 0.164 $\pm$ 0.10 \\
SkipthoughtDRAW & 2 & 11.2 & 18.9 & 63.3 & 0.157 $\pm$ 0.11 \\
noalignDRAW & 2.8 & 14.1 & 23.1 & 68 & 0.155 $\pm$ 0.11 \\
alignDRAW & 3 & 14 & 22.9 & 68.5 & 0.156 $\pm$ 0.11 \\
alignPixelRNN & 4.5 & 20.6 & 30.4 & 80.5 & 0.166 $\pm$ 0.12
\end{tabular}
\end{table}
The attention mechanism provides two key benefits to our model. Firstly, attention circumvents the model dependency over third-party segmentation or object detection by learning a latent alignment between text description and pixels. Hence, it enables the model to be learned in an end-to-end manner. Secondly, the attention enables us to decompose a single image generation problem into a series of smaller image generation problems rather than a single forward pass. This provides us a better understanding of how the model is generating the corresponding image portion. To the best of our knowledge, there are no reported results on the MNIST-with-captions dataset that we can compare our model with; therefore we only give our findings on the dataset in Table \ref{tab:2}.

\begin{table}[!h]
\centering
\caption{Results of proposed (i.e., PixelRNN) on MNIST-with-captions dataset.}
\label{tab:2}
\begin{tabular}{lccccc}
\hline
\multicolumn{1}{c}{\textbf{Model}} & \textbf{R@1} & \textbf{R@5} & \textbf{R@10} & \textbf{R@50} & \textbf{Image Similarity (SSI)} \\ \hline
PixelRNN (Mnist) & 4.3 & 53.6 & 63.1 & 88 & 0.356 $\pm$ 0.35
\end{tabular}
\end{table}

\section{Conclusion}
In this paper, we demonstrated that the autoregressive image modeling paradigm in combination with attention-based alignment model over words could generate images corresponding to given input captions. The attention-based model is introduced to remove the necessity of borrowing image segmentation or object detection models which results in an end-to-end learnable model. In comparison to the latent variable and generative adversarial network-based approaches, the proposed method has the advantages of being tractable and stable respectively. Nevertheless, the proposed model can be learned with adversarial training by employing it as a generator within GANs or GANs can be used as post-processor for further refining the results. A disadvantage of the proposed model is its computationally intensive training phase which is primarily due to sequential processing. In this regard, a future direction is to reduce the training period. For example, one possible approach is to impose constraints on pixel dependencies to allow parallel processing \cite{5}. As caption-based image generation modeling is relatively a novel area in image generation, the application area is still due to be developed.  

\bibliographystyle{unsrt}  
\bibliography{./references}

\begin{thebibliography}{10}

\bibitem{1}
Ruslan Salakhutdinov.
\newblock Learning deep generative models.
\newblock {\em Annual Review of Statistics and Its Application}, 2:361--385,
  2015.

\bibitem{2}
Alec Radford, Luke Metz, and Soumith Chintala.
\newblock Unsupervised representation learning with deep convolutional
  generative adversarial networks.
\newblock {\em arXiv preprint arXiv:1511.06434}, 2015.

\bibitem{3}
Ian Goodfellow, Jean Pouget-Abadie, Mehdi Mirza, Bing Xu, David Warde-Farley,
  Sherjil Ozair, Aaron Courville, and Yoshua Bengio.
\newblock Generative adversarial nets.
\newblock In {\em Advances in neural information processing systems}, pages
  2672--2680, 2014.

\bibitem{4}
Karol Gregor, Ivo Danihelka, Alex Graves, Danilo~Jimenez Rezende, and Daan
  Wierstra.
\newblock Draw: A recurrent neural network for image generation.
\newblock {\em arXiv preprint arXiv:1502.04623}, 2015.

\bibitem{5}
Aaron van~den Oord, Nal Kalchbrenner, and Koray Kavukcuoglu.
\newblock Pixel recurrent neural networks.
\newblock {\em arXiv preprint arXiv:1601.06759}, 2016.

\bibitem{6}
Lucas Theis and Matthias Bethge.
\newblock Generative image modeling using spatial lstms.
\newblock In {\em Advances in Neural Information Processing Systems}, pages
  1927--1935, 2015.

\bibitem{7}
Aaron Van~den Oord, Nal Kalchbrenner, Lasse Espeholt, Oriol Vinyals, Alex
  Graves, et~al.
\newblock Conditional image generation with pixelcnn decoders.
\newblock In {\em Advances in Neural Information Processing Systems}, pages
  4790--4798, 2016.

\bibitem{8}
Jon Gauthier.
\newblock Conditional generative adversarial nets for convolutional face
  generation.
\newblock {\em Class Project for Stanford CS231N: Convolutional Neural Networks
  for Visual Recognition, Winter semester}, 2014(5):2, 2014.

\bibitem{9}
Xinchen Yan, Jimei Yang, Kihyuk Sohn, and Honglak Lee.
\newblock Attribute2image: Conditional image generation from visual attributes.
\newblock In {\em European Conference on Computer Vision}, pages 776--791.
  Springer, 2016.

\bibitem{10}
Elman Mansimov, Emilio Parisotto, Jimmy~Lei Ba, and Ruslan Salakhutdinov.
\newblock Generating images from captions with attention.
\newblock {\em arXiv preprint arXiv:1511.02793}, 2015.

\bibitem{11}
Scott Reed, A{\"a}ron van~den Oord, Nal Kalchbrenner, Victor Bapst, Matt
  Botvinick, and Nando de~Freitas.
\newblock Generating interpretable images with controllable structure.
\newblock 2016.

\bibitem{12}
Junhyuk Oh, Xiaoxiao Guo, Honglak Lee, Richard~L Lewis, and Satinder Singh.
\newblock Action-conditional video prediction using deep networks in atari
  games.
\newblock In {\em Advances in Neural Information Processing Systems}, pages
  2863--2871, 2015.

\bibitem{13}
Leon~A Gatys, Alexander~S Ecker, and Matthias Bethge.
\newblock A neural algorithm of artistic style.
\newblock {\em arXiv preprint arXiv:1508.06576}, 2015.

\bibitem{14}
Kelvin Xu, Jimmy Ba, Ryan Kiros, Kyunghyun Cho, Aaron Courville, Ruslan
  Salakhudinov, Rich Zemel, and Yoshua Bengio.
\newblock Show, attend and tell: Neural image caption generation with visual
  attention.
\newblock In {\em International conference on machine learning}, pages
  2048--2057, 2015.

\bibitem{15}
Oriol Vinyals, Alexander Toshev, Samy Bengio, and Dumitru Erhan.
\newblock Show and tell: A neural image caption generator.
\newblock In {\em Proceedings of the IEEE conference on computer vision and
  pattern recognition}, pages 3156--3164, 2015.

\bibitem{16}
Andrej Karpathy and Li~Fei-Fei.
\newblock Deep visual-semantic alignments for generating image descriptions.
\newblock In {\em Proceedings of the IEEE conference on computer vision and
  pattern recognition}, pages 3128--3137, 2015.

\bibitem{17}
Alex Graves.
\newblock Generating sequences with recurrent neural networks.
\newblock {\em arXiv preprint arXiv:1308.0850}, 2013.

\bibitem{18}
Scott Reed, Zeynep Akata, Xinchen Yan, Lajanugen Logeswaran, Bernt Schiele, and
  Honglak Lee.
\newblock Generative adversarial text to image synthesis.
\newblock {\em arXiv preprint arXiv:1605.05396}, 2016.

\bibitem{19}
Diederik~P Kingma and Max Welling.
\newblock Auto-encoding variational bayes.
\newblock {\em arXiv preprint arXiv:1312.6114}, 2013.

\bibitem{20}
Danilo~Jimenez Rezende, Shakir Mohamed, and Daan Wierstra.
\newblock Stochastic backpropagation and approximate inference in deep
  generative models.
\newblock {\em arXiv preprint arXiv:1401.4082}, 2014.

\bibitem{21}
Alex Krizhevsky, Ilya Sutskever, and Geoffrey~E Hinton.
\newblock Imagenet classification with deep convolutional neural networks.
\newblock In {\em Advances in Neural Information Processing Systems}, pages
  1097--1105, 2012.

\bibitem{22}
Alex Graves, Navdeep Jaitly, and Abdel-rahman Mohamed.
\newblock Hybrid speech recognition with deep bidirectional lstm.
\newblock In {\em 2013 IEEE workshop on automatic speech recognition and
  understanding}, pages 273--278. IEEE, 2013.

\bibitem{23}
Dzmitry Bahdanau, Kyunghyun Cho, and Yoshua Bengio.
\newblock Neural machine translation by jointly learning to align and
  translate.
\newblock {\em arXiv preprint arXiv:1409.0473}, 2014.

\bibitem{24}
Geoffrey~E Hinton, Simon Osindero, and Yee-Whye Teh.
\newblock A fast learning algorithm for deep belief nets.
\newblock {\em Neural computation}, 18(7):1527--1554, 2006.

\bibitem{25}
Ruslan Salakhutdinov and Hugo Larochelle.
\newblock Efficient learning of deep boltzmann machines.
\newblock In {\em Proceedings of the thirteenth international conference on
  artificial intelligence and statistics}, pages 693--700, 2010.

\bibitem{26}
Hugo Larochelle and Iain Murray.
\newblock The neural autoregressive distribution estimator.
\newblock In {\em Proceedings of the Fourteenth International Conference on
  Artificial Intelligence and Statistics}, pages 29--37, 2011.

\bibitem{27}
Yoshua Bengio and Samy Bengio.
\newblock Modeling high-dimensional discrete data with multi-layer neural
  networks.
\newblock In {\em Advances in Neural Information Processing Systems}, pages
  400--406, 2000.

\bibitem{28}
Sepp Hochreiter and J{\"u}rgen Schmidhuber.
\newblock Long short-term memory.
\newblock {\em Neural computation}, 9(8):1735--1780, 1997.

\bibitem{29}
Scott Reed, A{\"a}ron van~den Oord, Nal Kalchbrenner, Sergio~G{\'o}mez
  Colmenarejo, Ziyu Wang, Yutian Chen, Dan Belov, and Nando de~Freitas.
\newblock Parallel multiscale autoregressive density estimation.
\newblock In {\em Proceedings of the 34th International Conference on Machine
  Learning-Volume 70}, pages 2912--2921. JMLR. org, 2017.

\bibitem{30}
Mehdi Mirza and Simon Osindero.
\newblock Conditional generative adversarial nets.
\newblock {\em arXiv preprint arXiv:1411.1784}, 2014.

\bibitem{31}
Junyoung Chung, Caglar Gulcehre, KyungHyun Cho, and Yoshua Bengio.
\newblock Empirical evaluation of gated recurrent neural networks on sequence
  modeling.
\newblock {\em arXiv preprint arXiv:1412.3555}, 2014.

\bibitem{32}
Tsung-Yi Lin, Michael Maire, Serge Belongie, James Hays, Pietro Perona, Deva
  Ramanan, Piotr Doll{\'a}r, and C~Lawrence Zitnick.
\newblock Microsoft coco: Common objects in context.
\newblock In {\em European conference on computer vision}, pages 740--755.
  Springer, 2014.

\bibitem{33}
Mart{\'\i}n Abadi, Paul Barham, Jianmin Chen, Zhifeng Chen, Andy Davis, Jeffrey
  Dean, Matthieu Devin, Sanjay Ghemawat, Geoffrey Irving, Michael Isard, et~al.
\newblock Tensorflow: A system for large-scale machine learning.
\newblock In {\em 12th $\{$USENIX$\}$ Symposium on Operating Systems Design and
  Implementation ($\{$OSDI$\}$ 16)}, pages 265--283, 2016.

\bibitem{34}
Zhou Wang, Alan~C Bovik, Hamid~R Sheikh, Eero~P Simoncelli, et~al.
\newblock Image quality assessment: from error visibility to structural
  similarity.
\newblock {\em IEEE transactions on image processing}, 13(4):600--612, 2004.

\bibitem{35}
Ryan Kiros, Yukun Zhu, Ruslan~R Salakhutdinov, Richard Zemel, Raquel Urtasun,
  Antonio Torralba, and Sanja Fidler.
\newblock Skip-thought vectors.
\newblock In {\em Advances in neural information processing systems}, pages
  3294--3302, 2015.

\end{thebibliography}

\end{document}